\documentclass[runningheads]{llncs}

\usepackage{amsmath,amssymb} 
\usepackage{times}
\usepackage{epsfig}
\usepackage{graphicx}
\usepackage{algorithm}
\usepackage{multirow}
\usepackage{mathtools,xparse}
\usepackage{subcaption}
\usepackage{siunitx}
\usepackage{stmaryrd}
\usepackage[noend]{algpseudocode}
\usepackage{placeins}
\usepackage[first=0,last=9]{lcg}
\usepackage{mdframed}
\usepackage{color, colortbl}
\usepackage[table,dvipsnames]{xcolor}

\algnewcommand\algorithmicforeach{\textbf{for each}}
\algdef{S}[FOR]{ForEach}[1]{\algorithmicforeach\ #1\ \algorithmicdo}

\newcommand{\trans}{\text{T}}
\newcommand{\matr}[1]{\mathbf{#1}}     

\newcommand{\Hom}{\matr{H}}
\newcommand{\Aff}{\matr{A}}
\newcommand{\Fund}{\matr{F}}
\newcommand{\Ess}{\matr{E}}

\newcommand{\Point}{\matr{p}}
\newcommand*\rot{\rotatebox{90}}

\DeclarePairedDelimiter{\norm}{\lVert}{\rVert}
\NewDocumentCommand{\normL}{ s O{} m }{%
  \IfBooleanTF{#1}{\norm*{#3}}{\norm[#2]{#3}}_{2}%
}

\newcommand{\mc}[3]{\multicolumn{#1}{#2}{#3}}
\newcommand{\ph}[1]{\phantom{#1}}
\definecolor{Gray}{gray}{0.85}
\newcolumntype{g}{>{\columncolor{Gray}}c}

\usepackage[pagebackref=true,breaklinks=true,letterpaper=true,colorlinks,bookmarks=false]{hyperref}

\begin{document}
\pagestyle{headings}
\mainmatter

\def\ACCV18SubNumber{43}  

\title{Recovering affine features from \\orientation- and scale-invariant ones} 
\titlerunning{Recovering affine features from orientation- and scale-invariant ones}
\authorrunning{Daniel Barath}

\author{Daniel Barath$^{12}$}
\institute{$^1$ Centre for Machine Perception, Czech Technical University, Prague, Czech Republic\\
$^2$ Machine Perception Research Laboratory, MTA SZTAKI, Budapest, Hungary}

\maketitle

\begin{abstract}
An approach is proposed for recovering affine correspondences (ACs) from orientation- and scale-invariant, e.g.\ SIFT, features. 
The method calculates the affine parameters consistent with a pre-estimated epipolar geometry from the point coordinates and the scales and rotations which the feature detector obtains.
The closed-form solution is given as the roots of a quadratic polynomial equation, thus having two possible real candidates and fast procedure, i.e.\ $<1$ millisecond.
It is shown, as a possible application, that using the proposed algorithm allows us to estimate a homography for every single correspondence independently.
It is validated both in our synthetic environment and on publicly available real world datasets, that the proposed technique leads to accurate ACs. Also, the estimated homographies have similar accuracy to what the state-of-the-art methods obtain, but due to requiring only a single correspondence, the robust estimation, e.g.\ by locally optimized RANSAC, is an order of magnitude faster. 
\end{abstract}

\section{Introduction}

This paper addresses the problem of recovering fully affine-covariant features \cite{mikolajczyk2005comparison} from orientation- and scale-invariant ones obtained by, for instance, SIFT \cite{lowe1999object} or SURF \cite{bay2006surf} detectors. This objective is achieved by considering the epipolar geometry to be known between two images and exploiting the geometric constraints which it implies.\footnote{Note that the pre-estimation of the epipolar geometry, either that of a fundamental $\Fund$ or essential matrix $\Ess$, is usual in computer vision applications.} 
The proposed algorithm requires the epipolar geometry, i.e.\ characterized by either a fundamental $\Fund$ or an essential $\Ess$ matrix, and an orientation and scale-invariant feature as input and returns the affine correspondence consistent with the epipolar geometry.

Nowadays, a number of solutions is available for estimating geometric models from affine-covariant features. 
For instance, Perdoch et al.~\cite{PerdochMC06} proposed techniques for approximating the epipolar geometry between two images by generating point correspondences from the affine features. 
Bentolila and Francos~\cite{Bentolila2014} showed a method to estimate the exact, i.e.\ with no approximation, $\Fund$ from three correspondences. Raposo et al.~\cite{Raposo2016} proposed a solution for essential matrix estimation using two feature pairs. Bar\'ath et al.~\cite{barath2017focal} proved that even the semi-calibrated case, i.e.\ when the objective is to find the essential matrix and a common focal length, is solvable from two correspondences. 
Homographies can also be estimated from two features~\cite{koser2009geometric} without any a priori knowledge about the camera movement. In case of known epipolar geometry, a single affine correpondence is enough for estimating a homography~\cite{barath2017theory}. 
Also, local affine transformations encode the surface normals~\cite{Molnar2014}. Therefore, if the cameras are calibrated, the normal can be unambiguously estimated from a single affine correspondence~\cite{koser2009geometric}. 
Pritts et al.~\cite{Pritts2017RadiallyDistortedCT} recently showed that the lens distortion parameters can also be retrieved. 

Affine correspondences encode higher-order information about the underlying scene geometry and this is what makes the listed algorithms able to estimate geometric models, e.g.\ homographies and fundamental matrices, exploiting only a few correspondences -- significantly less than what point-based methods require. This however implies the major drawback of the previously mentioned techniques: obtaining affine correspondences accurately (for example, by applying Affine-SIFT \cite{morel2009asift}, MODS \cite{mishkin2015mods}, Hessian-Affine, or Harris-Affine \cite{mikolajczyk2005comparison} detectors) in real image pairs is time consuming and thus, these methods are not applicable when real time performance is necessary. 
In this paper, the objective of the proposed method is to bridge this problem by recovering the full affine correspondence from only a part of it, i.e.\ the feature rotation and scale. This assumption is realistic since some of the widely-used feature detectors, e.g.\ SIFT or SURF, return these parameters besides the point coordinates. 

Interestingly, the exploitation of these additional affine components is not often done in geometric model estimation applications, even though, it is available without demanding additional computation.
Using only a part of an affine correspondence, e.g.\ solely the rotation component, is a well-known technique, for instance, in wide-baseline feature matching~\cite{matas2004robust,mishkin2015mods}. However, to the best of our knowledge, there are only two papers~\cite{barath2017phaf,barath2018five} involving them to geometric model estimation. In \cite{barath2017phaf}, $\Fund$ is assumed to be known a priori and a technique is proposed for estimating a homography using two SIFT correspondences exploiting their scale and rotation components. Even so, an assumption is made, considering that the scales along the horizontal and vertical axes are equal to that of the SIFT features -- which generally does not hold. Thus, the method yields only an approximation. The method of \cite{barath2018five} obtains the fundamental matrix by first estimating a homography using three point correspondences and the feature rotations. Then $\Fund$ is retrieved from the homography and two additional correspondences. 

The contributions of the paper are: (i) we propose a technique for estimating affine correspondences from orientation- and scale-invariant features in case of known epipolar geometry.\footnote{Note that the proposed method can straightforwardly be generalized for multiple fundamental matrices, i.e.\ multiple rigid motions.} The method is fast, i.e.\ $<1$ milliseconds, due to being solved in closed-form as the roots of a quadratic polynomial equation. (ii) It is validated both in our synthetic environment and on more than $9000$ publicly available real image pairs that homography estimation is possible from a single SIFT correspondence accurately. Benefiting from the number of correspondence required, robust estimation, by e.g.\ LO-RANSAC~\cite{chum2003locally}, is an order of magnitude faster than by combining it with the standard techniques, e.g.\ four- and three-point algorithms~\cite{hartley2003multiple}.    

\section{Theoretical Background}

\paragraph{Affine Correspondences.}

In this paper, we consider an affine correspondence (AC) as a triplet: $(\Point_1, \Point_2, \Aff)$, where $\Point_1 = [u_1 \quad v_1 \quad 1]^\trans$ and $\Point_2 = [u_2 \quad v_2 \quad 1]^\trans$ are a corresponding homogeneous point pair in the two images (the projections of point $\textbf{P}$ in Fig.~\ref{fig:motion}), and 
\begin{equation*}
	\Aff = \begin{bmatrix}
		a_{1} & a_{2} \\
		a_{3} & a_{4}
	\end{bmatrix}
\end{equation*}
is a $2 \times 2$ linear transformation which we call \textit{local affine transformation}. To define $\Aff$, we use the definition provided in~\cite{Molnar2014} as it is given as the first-order Taylor-approximation of the $\text{3D} \to \text{2D}$ projection functions. Note that, for perspective cameras, the formula for $\Aff$ simplifies to the first-order approximation of the related \textit{homography} matrix
\begin{equation*}
	\Hom = \begin{bmatrix}
		h_1 & h_2 & h_3 \\
		h_4 & h_5 & h_6 \\
		h_7 & h_8 & h_9
	\end{bmatrix}
\end{equation*}
as follows: 
\begin{eqnarray}
		\begin{array}{lllllll}
          a_{1} & = & \frac{\partial u_2}{\partial u_1} = \frac{h_{1} - h_{7} u_2}{s}, & \phantom{xxxx} & 
          a_{2} & = & \frac{\partial u_2}{\partial v_1} = \frac{h_{2} - h_{8} u_2}{s}, \\[2mm]
          a_{3} & = & \frac{\partial v_2}{\partial u_1} = \frac{h_{4} - h_{7} v_2}{s}, & &
          a_{4} & = & \frac{\partial v_2}{\partial v_1} = \frac{h_{5} - h_{8} v_2}{s}, 
		\end{array}
        \label{eq:taylor_approximation}
\end{eqnarray}
where $u_i$ and $v_i$ are the directions in the $i$th image ($i \in \{1,2\}$) and $s = u_1 h_7 + v_1 h_8 + h_9$ is the so-called projective depth.

\paragraph{Fundamental matrix} 
\begin{equation*}
	\Fund = \begin{bmatrix}
		f_1 & f_2 & f_3 \\
		f_4 & f_5 & f_6 \\
		f_7 & f_8 & f_9
	\end{bmatrix}
\end{equation*}
is a $3 \times 3$ transformation matrix ensuring the so-called epipolar constraint $\Point_2^\trans \matr{F} \Point_1 = 0$ for rigid scenes. Since its scale is arbitrary and $\det(\Fund) = 0$, matrix $\Fund$ has seven degrees-of-freedom (DoF). 
The geometric relationship of $\Fund$ and $\Aff$ was first shown in \cite{barath2017focal} and interpreted by formula $\Aff^{-\trans} (\Fund^\trans \Point_2)_{(1:2)} + (\Fund \Point_1)_{(1:2)} = 0$, where lower index $(i:j)$ selects the sub-vector consisting of the elements from the $i$th to the $j$th. 
These properties will help us to recover the full affine transformation from a SIFT correspondence. 

\section{Recovering affine correspondences}

In this section, we show how can affine correspondences be recovered from rotation- and scale-invariant features in case of known epipolar geometry. 
Even though we will use SIFT as an alias for this kind of features, the derived formulas hold for the output of every scale- and orientation-invariant detector.
First, the affine transformation model is described in order to interpret the SIFT angles and scales. Then this model is substituted into the relationship of affine transformations and fundamental matrices. Finally, the obtained system is solved in closed-form to recover the unknown affine parameters. 

\subsection{Affine transformation model}

\begin{figure}[htb]
	\centering
    \begin{subfigure}[t]{0.49\textwidth}
    	\includegraphics[width = 0.95\columnwidth]{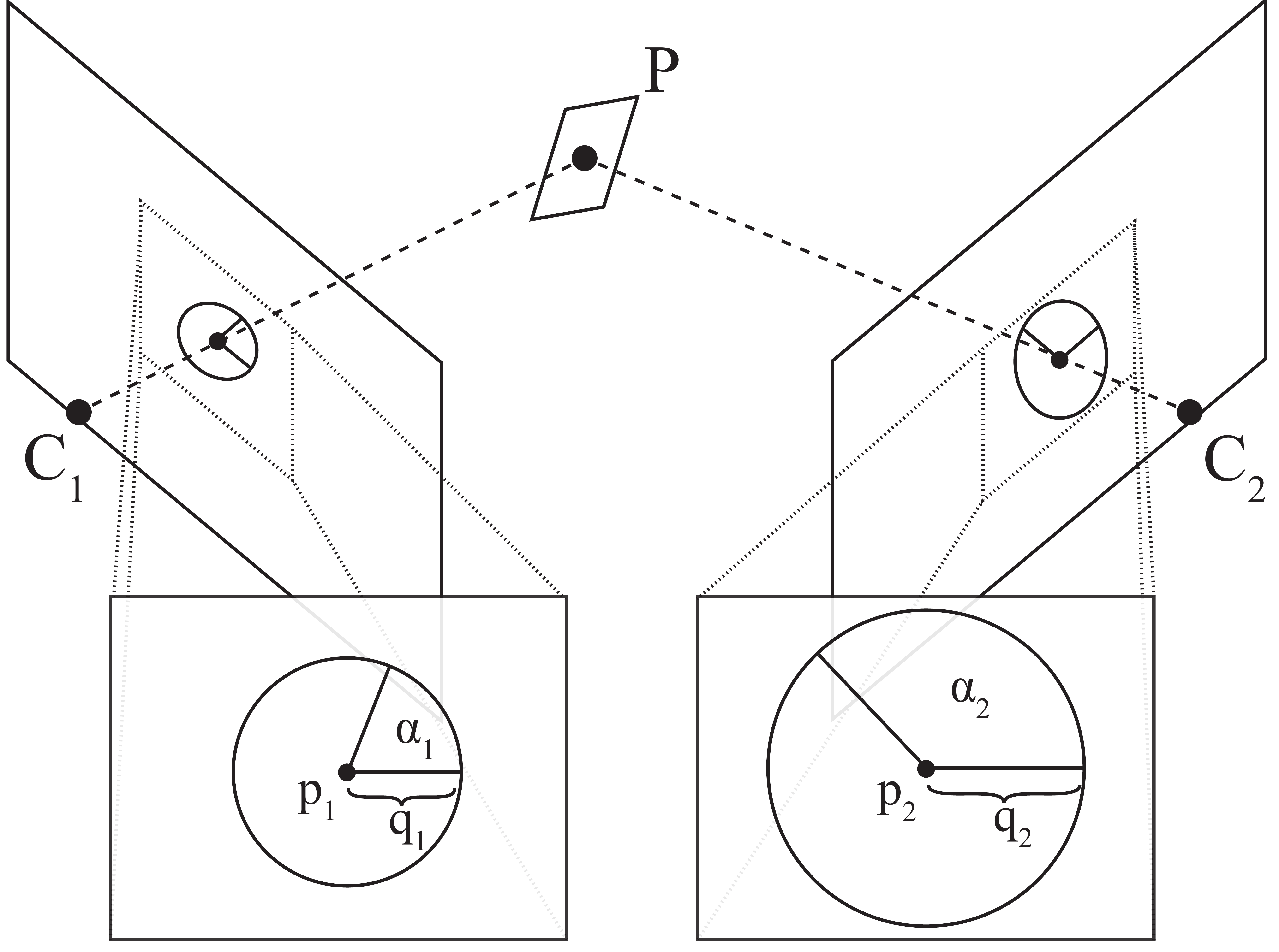}
    	\caption{ Point $\textbf{P}$ and the surrounding patch projected into cameras $\textbf{C}_1$ and $\textbf{C}_2$. A window showing the projected points $\textbf{p}_1 = [u_1 \; v_1 \; 1]^\trans$ and $\textbf{p}_2 = [u_2 \; v_2 \; 1]^\trans$ are cut out and enlarged. The rotation of the feature in the $i$th image is $\alpha_i$ and the size is $q_i$ ($i \in \{1,2\}$). The scaling from the $1$st to the $2$nd image is calculated as $q = q_2 / q_1$. }
    	\label{fig:motion}
    \end{subfigure}\hfill
    \begin{subfigure}[t]{0.49\textwidth}
   	 	\includegraphics[width = 0.95\columnwidth]{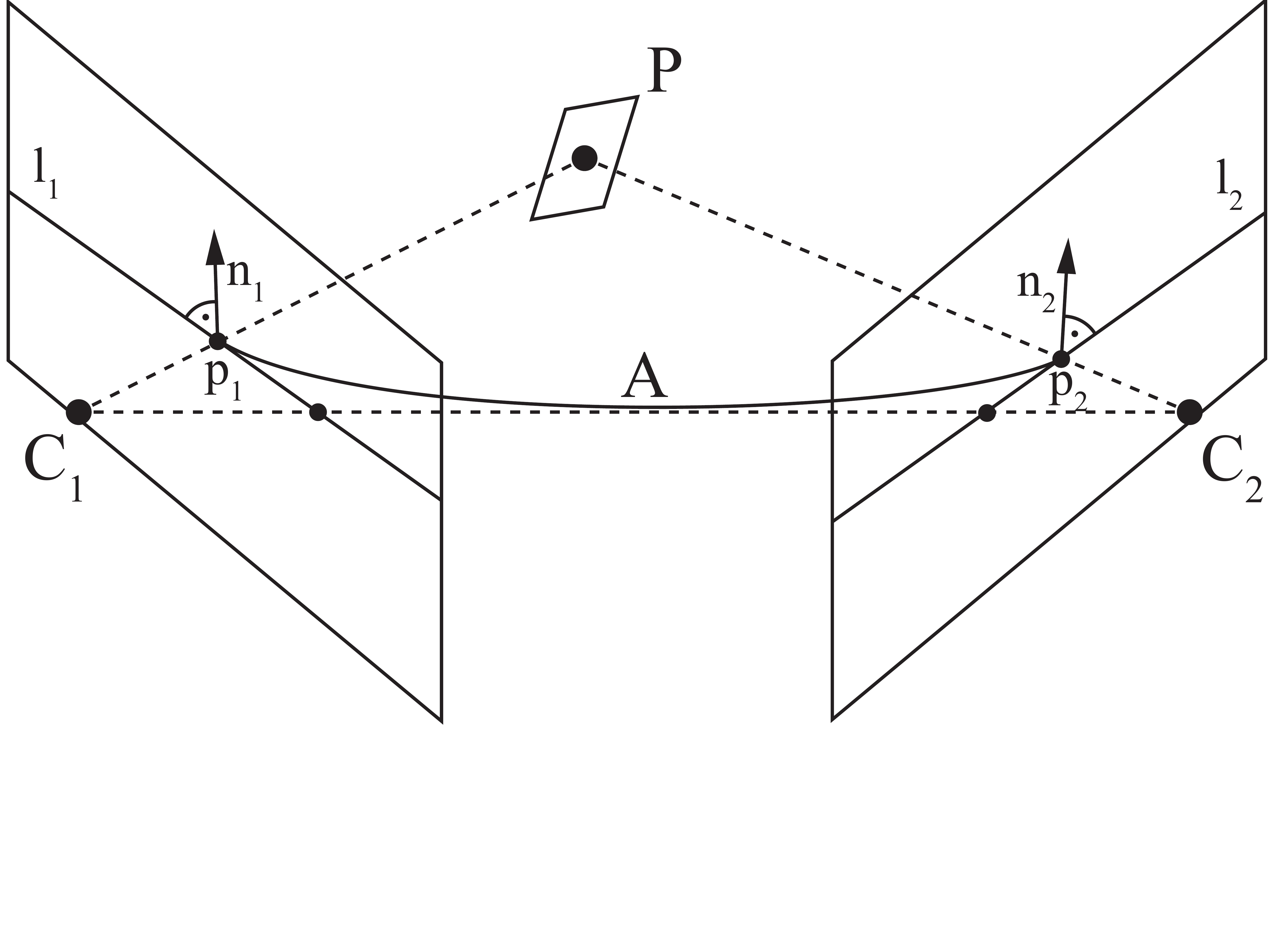}
    	\caption{ The geometric interpretation of the relationship of a local affine transformations and the epipolar geometry (Eq.~\ref{eq:rotating_normals}; proposed in \cite{barath2017focal}). Given the projection $\matr{p}_i$ of $\matr{P}$ in the $i$th camera $\matr{C}_i$, $i \in \{1,2\}$. The normal $\matr{n}_1$ of epipolar line $\matr{l}_1$ is mapped by affinity $\Aff \in \mathbb{R}^{2 \times 2}$ into the normal $\matr{n}_2$ of epipolar line $\matr{l}_2$.}
    	\label{fig:rotating_normals}
    \end{subfigure}
    \caption{ (a) The geometric interpretation of orientation- and scale-invariant features. (b) The relationship of local affine transformation and epipolar geometry. }
\end{figure}

The objective of this section is to define an affine transformation model which interprets the scale and rotation components of the features. Suppose that we are given a rotation $\alpha_i$ and scale $q_i$ in each image ($i \in \{1, 2\}$) besides the point coordinates. For the geometric interpretation of the features, see Fig.~\ref{fig:motion}. 
Reflecting the fact that the two rotations act on different images, we interpret $\Aff$ as follows: 
\begin{eqnarray*}
	\Aff = \matr{R}_{\alpha_2} \matr{U} \matr{R}_{-\alpha_1} = \\[2mm]
    \begin{bmatrix}
    	a_1 & a_2 \\
        a_3 & a_4
    \end{bmatrix} = 
    \begin{bmatrix}
    	c_2 & -s_2\\
        s_2 & c_2
    \end{bmatrix} 
    \begin{bmatrix}
    	q_u & w \\
        0 & q_v
    \end{bmatrix} 
    \begin{bmatrix}
    	c_1 & -s_1 \\
        s_1 & c_1
    \end{bmatrix},
\end{eqnarray*}

\noindent
where $\alpha_1$ and $\alpha_2$ are the two rotations in the images obtained by the feature detector; $q_u$ and $q_v$ are the scales along the horizontal and vertical axes; $w$ is the shear parameter; and $c_1 = \cos(-\alpha_1)$, $s_1 = \sin(-\alpha_1)$, $c_2 = \cos(\alpha_2)$, $s_2 = \sin(\alpha_2)$. Thus $\Aff$ is written as a multiplication of two rotations ($\matr{R}_{-\alpha_1}$ and $\matr{R}_{\alpha_2}$) and an upper triangle matrix $\matr{U}$. Since a uniform scale $q_i$ is given for each image, we calculate the scale between the images as $q = q_2 / q_1$. Even though $q$ is thus considered as a known parameter, $q_u$ and $q_v$ are still unknowns. However, the scales obtained by scale-invariant feature detectors are calculated from the sizes of the corresponding regions. Therefore, it can be easily seen that,
\begin{equation}
	q = \det \Aff = q_u q_v.
    \label{eq:ScaleDependency}
\end{equation}
After the multiplication of the matrices, the affine elements are as follows:
\begin{equation}
	\begin{array}{c c c}
      a_1 & = & c_1 c_2 q_u - s_1 s_2 q_v + c_1 s_2 w, \\[1mm]
      a_2 & = & -c_1 s_2 q_u - s_1 c_2 q_v + c_1 c_2 w,  \\[1mm]
      a_3 & = & s_1 c_2 q_u + c_1 s_2 q_v + s_1 s_2 w, \\[1mm]
      a_4 & = & -s_1 s_2 q_u + c_1 c_2 q_v + s_1 c_2 w. 
	\end{array}
    \label{eq:DecomposedParameters}
\end{equation}
Using these equations, the rotations and scales which the rotation- and scale-invariant feature detectors obtain are interpretable in terms of perspective geometry. 
Note that this decomposition is not unique, as it is discussed in the appendix. However, in this way, the rotations are interpreted affecting independently in the two images. 

Also note that \cite{barath2017phaf} also proposes a decomposition for $\Aff$, however, their method assumes that the rotation equals to $\alpha_2 - \alpha_1$ and therefore does not reflect the fact that feature detectors obtain these affine components separately in the images. Also, due to assuming that $q_u = q_v$, the algorithm of \cite{barath2017phaf} obtains only an approximation -- the error is not zero even in the noise-free case. 

\subsection{Affine correspondence from epipolar geometry}

Estimating the epipolar geometry as a preliminary step, either a fundamental or an essential matrix, is often done in computer vision applications. In the rest of the paper, we consider fundamental matrix $\Fund$ to be known in order to exploit the relationship of epipolar geometry and affine correspondences proposed in~\cite{barath2017focal}. Note that the formulas proposed in this paper hold for essential matrix $\Ess$ as well. 

For a local affine transformation $\Aff$ consistent with $\matr{F}$, formula
\begin{equation}
	\label{eq:rotating_normals}
	\Aff^{-\trans} \matr{n}_1 = -\matr{n}_2,
\end{equation}
holds, where $\matr{n}_1$ and $\matr{n}_2$ are the normals of the epipolar lines in the first and second images regarding to the observed point locations (see Fig.~\ref{fig:rotating_normals}). These normals are calculated as follows: $\matr{n}_1 = (\matr{F}^\trans \matr{p}_2)_{(1:2)}$ and $\matr{n}_2 = (\matr{F} \matr{p}_1)_{(1:2)}$, where lower index $({1:2})$ selects the first two elements of the input vector. This relationship can be written by a linear equation system consisting of two equations, one for each coordinate of the normals, as 
\begin{eqnarray}
	\begin{array}{r c c}
	(u_2 + a_1 u_1) f_1 + a_1 v_1 f_2 + a_1 f_3 + 
    (v_2 + a_3 u_1) f_4 + a_3 v_1 f_5 + a_3 f_6 + f_7 & = & 0, \\[2mm]
	a_2 u_1 f_1 + (u_2 + a_2 v_1) f_2 + a_2 f_3 + a_4 u_1 f_4 +
    (v_2 + a_4 v_1) f_5 + a_4 f_6 + f_8 & = & 0,
	\end{array}
    \label{eq:FA12}
\end{eqnarray}
where $f_i$ is the $i$th ($i \in [1, 9])$ element of the fundamental matrix in row-major order, $a_j$ is the $j$th element of the affine transformation and each $u_k$ and $v_k$ are the point coordinates in the $k$th image ($k \in \{ 1, 2 \}$).
Assuming that $\matr{F}$ and point coordinates $(u_1, v_1)$, $(u_2, v_2)$ are known and the only unknowns are the affine parameters, Eqs.~\ref{eq:FA12} are reformulated as follows:
\begin{eqnarray}
	\begin{array}{r}
	(u_1 f_1 + v_1 f_2 + f_3) a_1 + (u_1 f_4 + v_1 f_5 + f_6) a_3 = 
    -u_2 f_1 - v_2 f_4 - f_7, \\[2mm] 
	(u_1 f_1 + v_1 f_2 + f_3) a_2 + (u_1 f_4 + v_1 f_5 + f_6) a_4 = 
    -u_2 f_2 - v_2 f_5 - f_8.
	\end{array}
    \label{eq:FA2}
\end{eqnarray}
These equations are linear in the affine components. Let us replace the constant parameters by variables and thus introduce the following notation:
\begin{eqnarray*}
	B = u_1 f_1 + v_1 f_2 + f_3, & \phantom{xxxx} & C = u_1 f_4 + v_1 f_5 + f_6, \\
    D = -u_2 f_1 - v_2 f_4 - f_7, & & E = -u_2 f_2 - v_2 f_5 - f_8.
\end{eqnarray*}
Therefore, Eqs.~\ref{eq:FA2} become
\begin{eqnarray}
	\begin{array}{r c r}
	B a_1 + C a_3 = D, & \phantom{xxxx} & B a_2 + C a_4 = E.
	\end{array}
    \label{eq:FA3}
\end{eqnarray}
By substituting Eqs.~\ref{eq:DecomposedParameters} into Eqs.~\ref{eq:FA3} the following formula is obtained:
\begin{eqnarray}
	\begin{array}{r c r}
	B (c_1 c_2 q_u - s_1 s_2 q_v + c_1 s_2 w) + C (s_1 c_2 q_u + c_1 s_2 q_v + s_1 s_2 w) & = & D, \\[2mm]
    B (-c_1 s_2 q_u - s_1 c_2 q_v + c_1 c_2 w) + C (-s_1 s_2 q_u + c_1 c_2 q_v + s_1 c_2 w) & = & E.
	\end{array}
    \label{eq:FA6}
\end{eqnarray}
Since the rotations, and therefore their sinuses ($s_1$, $s_2$) and cosines ($c_1$, $c_2$), are considered to be known, Eqs.~\ref{eq:FA6} are re-arranged as follows: 
\begin{equation}
	\begin{array}{r c c}
	(B c_1 c_2 + C s_1 c_2) q_u + (C c_1 s_2 - B s_1 s_2) q_v +
    (B c_1 s_2 + C s_1 s_2) w & = & D, \\[2mm] 
	(-B c_1 s_2 - C s_1 s_2) q_u + (C c_1 c_2 - B s_1 c_2) q_v +
    (B c_1 c_2 + C s_1 c_2) w & = & E.
	\end{array}	
    \label{eq:FA5}
\end{equation}
Let us introduce new variables encapsulating the constants as 
\begin{eqnarray*}
	G = B c_1 c_2 + C s_1 c_2, & \phantom{xxxx} & H = C c_1 s_2 - B s_1 s_2, \\
    I = B c_1 s_2 + C s_1 s_2, & & J = -B c_1 s_2 - C s_1 s_2. \\
    K = C c_1 c_2 - B s_1 c_2, & & 
\end{eqnarray*}
Eqs.~\ref{eq:FA5} are then become
\begin{equation}
	\begin{array}{r c c}
      G q_u + H q_v + I w & = & D, \\[2mm] 
      J q_u + K q_v + G w  & = & E.
	\end{array}	
    \label{eq:FA4}
\end{equation}
From the first equation, we express $w$ as follows:
\begin{equation}
	w = \frac{D}{I} - \frac{G}{I} q_u - \frac{H}{I} q_v.
    \label{eq:w}
\end{equation}
Let us notice that $q_u$ and $q_v$ are dependent due to Eq.~\ref{eq:ScaleDependency} as $q_u = q / q_v$. By substituting this formula and Eq.~\ref{eq:w} into the second equation of Eqs.~\ref{eq:FA4}, the following quadratic polynomial equation is given:
\begin{equation*}
      \left(K + \frac{G D - G H}{I} \right) q_v^2 - \left(\frac{G^2 q}{I} + E \right) q_v + J q = 0.
\end{equation*}
After solving the polynomial equation which has two solutions $q_{v,1}$ and $q_{v,2}$, all the other parameters of each solution can be straightforwardly calculated as
\begin{equation*}
	q_{u,i} = \frac{q}{q_{v,i}}, \quad\quad w_i = \frac{D}{I} - \frac{G}{I} q_{u,i} - \frac{H}{I} q_{v,i}, \quad\quad i \in \{1,2\}.
\end{equation*}
Consequently, each SIFT correspondence lead to two possible affine correspondences. Therefore, \textit{we recovered the local affine transformation from an orientation- and scale-invariant correspondence in case of known epipolar geometry}. Note that a good heuristics for rejecting invalid affinities is to discard those having extreme scaling or shearing. 
\begin{figure*}[htb]
	\centering
    \includegraphics[width=0.325\textwidth]{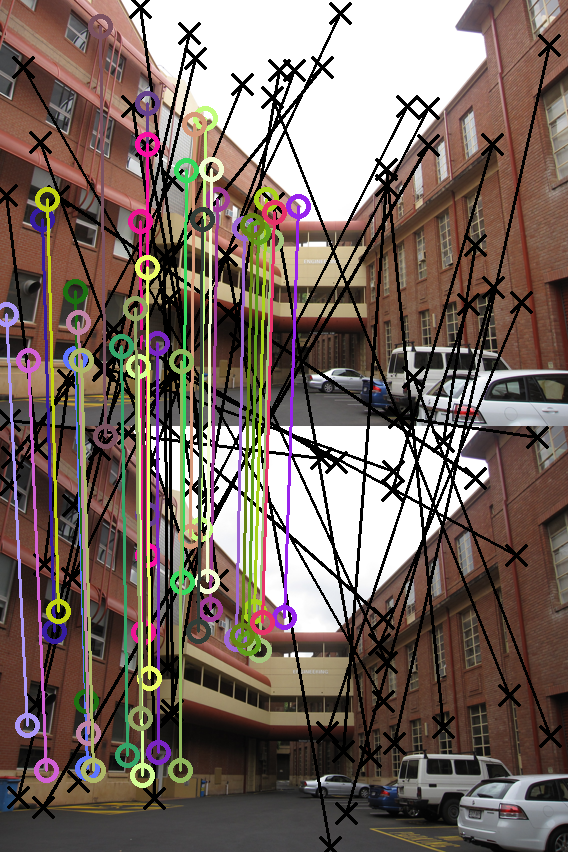}
    \includegraphics[width=0.325\textwidth]{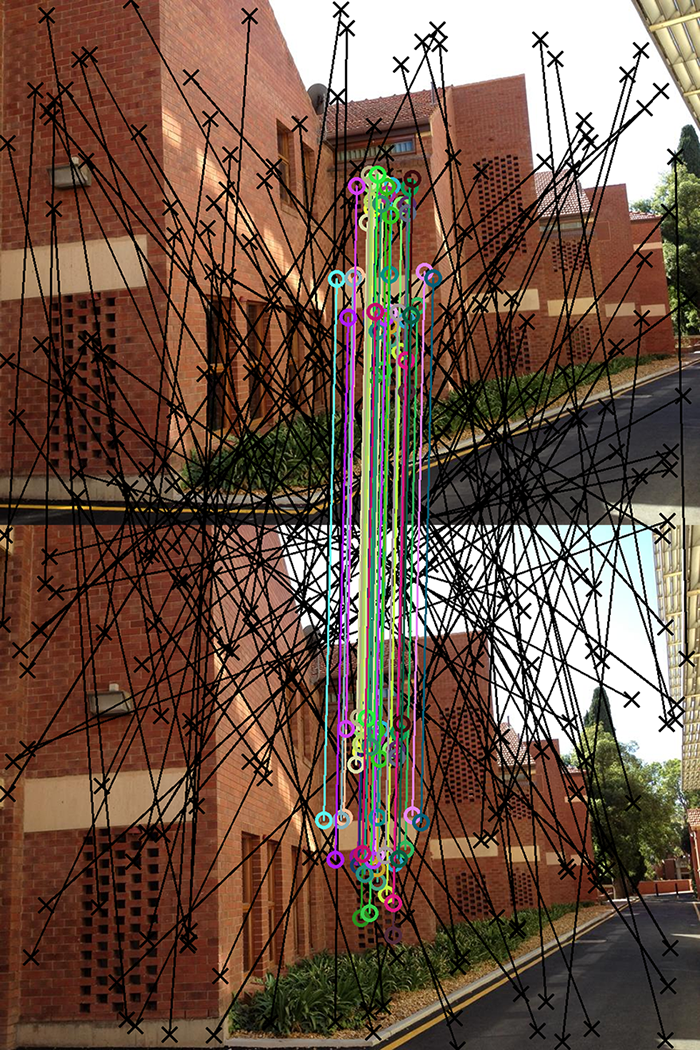}
    \includegraphics[width=0.325\textwidth]{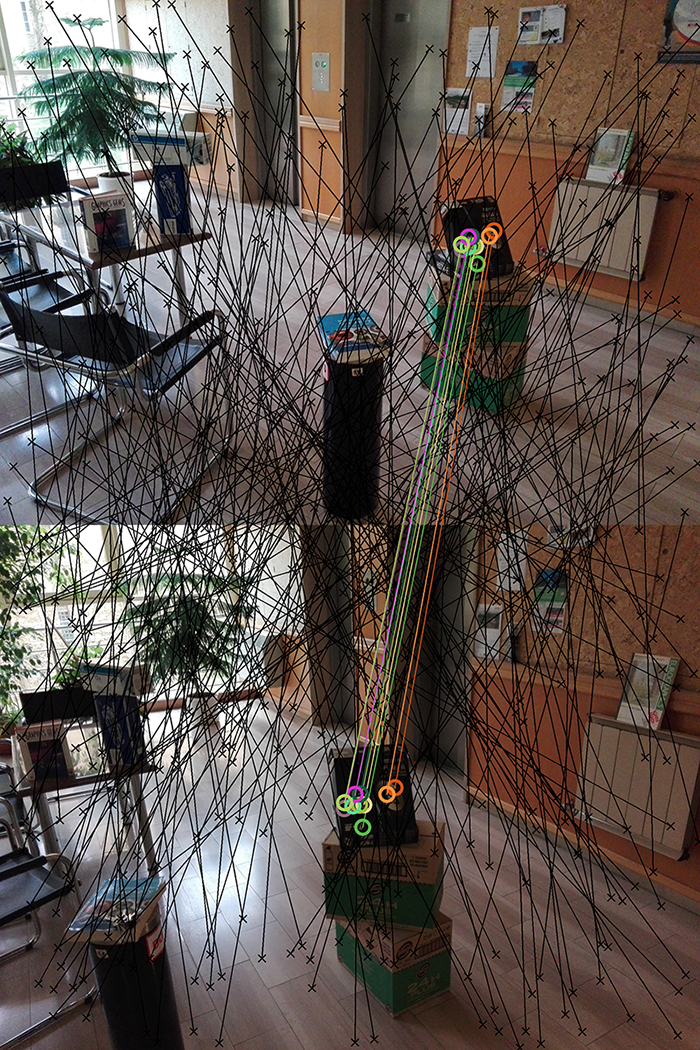}
    \caption{ Inlier (circles) and outlier (black crosses) correspondences found by the proposed method on image pairs from the {\fontfamily{cmtt}\selectfont AdelaideRMF} ($1$st and $2$nd columns) and {\fontfamily{cmtt}\selectfont Multi-H} datasets ($3$rd). Every $5$th correspondence is drawn. }
    \label{fig:datasets_1}
\end{figure*}

\section{Experimental results}

In this section, we compare the affine correspondences obtained by the proposed method with techniques approximating them. Then it is demonstrated that by using the affinities recovered from a SIFT correspondence, a homography can be estimated from a single correspondence. Due to requiring only a single correspondence, robust homography estimation becomes significantly faster, i.e.\ an order of magnitude, than by using the traditional techniques, for instance, the four- or three-point algorithms~\cite{hartley2003multiple}. 

\subsection{Comparing techniques to estimate affine correspondences}

For testing the accuracy of the affine correspondences obtained by the proposed method, first, we created a synthetic scene consisting of two cameras represented by their $3 \times 4$ projection matrices $\matr{P}_1$ and $\matr{P}_2$. They were located in random surface points of a 10-radius center-aligned sphere. A plane with random normal was generated in the origin and ten random points, lying on the plane, were projected into both cameras. To get the ground truth affine transformations, we first calculated homography $\Hom$ by projecting four random points from the plane to the cameras and applying the normalized direct linear transformation~\cite{hartley2003multiple} algorithm to them. The local affine transformation regarding to each correspondence were computed from the ground truth homography as its first order Taylor-approximation by Eq.~\ref{eq:taylor_approximation}. 
Note that $\Hom$ could have been calculated directly from the plane parameters as well. However, using four points promised an indirect but geometrically interpretable way of noising the affine parameters: by adding zero-mean Gaussian-noise to the coordinates of the four projected points which implied $\Hom$. 
Finally, after having the full affine correspondence, $\Aff$ was decomposed to $\matr{R}_{-\alpha}$, $\matr{R}_{\beta}$ and $\matr{U}$ in order to simulate the SIFT output. The decomposition is discussed in the appendix in depth. 
Since the decomposition is ambiguous, due to the two angles, $\beta$ was set to a random value.  
Zero-mean Gaussian noise was added to the point coordinates and the affine transformations were noised in the previously described way. 
The error of an estimated affinity is calculated as $| \Aff_\text{est} - \Aff_{gt} |_\text{F}$, where $\Aff_\text{est}$ is the estimated affine matrix, $\Aff_\text{gt}$ is the ground truth one and norm $| . |_{\text{F}}$ is the Frobenious-norm. Out of the at most two real solutions of the proposed algorithm, we selected the one which is the closest to the ground truth.

Figure~\ref{fig:affine_error_increasing_noise} reports the 
error of the estimated affinities plotted as the function of the noise $\sigma$.
The affine transformations were estimated by the proposed method (red curve), approximated as $\textbf{A} \approx \matr{R}_{\beta - \alpha} \textbf{D}$ (green; proposed in \cite{barath2017phaf}) and as $\textbf{A} \approx\matr{R}_{\beta} \textbf{D} \matr{R}_{-\alpha}$ (blue), where $\textbf{R}_\theta$ is 2D rotation matrix rotating by $\theta$ degrees and $\textbf{D} = \text{diag}(q, q)$. 
Note that $\textbf{D} \matr{R}_{\beta - \alpha}$ does not have to be tested since $\matr{R}_{\beta - \alpha} \textbf{D} = \textbf{D} \matr{R}_{\beta - \alpha}$.
To Figure~\ref{fig:affine_error_increasing_noise}, approximating the affine transformation by the tested ways lead to inaccurate affine estimates -- the error is not zero even in the noise-free case. The proposed method behaves reasonably as the noise $\sigma$ increases and the error is zero if there is no noise. 

\subsection{Application: homography estimation}

In~\cite{barath2017theory}, a method, called HAF, was published for estimating the homography from a single affine correspondence. The method requires the fundamental matrix and an affine correspondence to be known between two images. Assuming that $\textbf{P}$ is on a continuous surface, HAF estimates homography $\Hom$ which the tangent plane of the surface at point $\textbf{P}$ implies.
The solution is obtained by first exploiting the fundamental matrix and reducing the number of unknowns in $\Hom$ to four. Then the relationship written in Eq.~\ref{eq:taylor_approximation} is used to express the remaining homography parameters by the affine correspondences. The obtained inhomogeneous linear system consists of six equations for four unknowns. 
The problem to solve is $\matr{C} \matr{x} = \matr{b}$, where $\matr{x} = [h_7, \; h_8, \; h_9]^\trans$ is the vector of unknowns, i.e.\ the last row of $\Hom$,  vector $\matr{b} = [f_4, \; f_5, \; -f_1, \; -f_2, \; -u_1 f_4 - v_1 f_5 - f_6,  \; u_1 f_1 + v_1 f_2 - f_3]$ is the inhomogeneous part and $\matr{C}$ is the coefficient matrix as follows:
\begin{equation*}
	\matr{C} = \begin{bmatrix}
		a_1 u_1 + u_2 - e_u & a_1 v_1 & a_1 \\
        a_2 u_1 & a_2 v_1 + u_2 - e_u & a_2 \\
        a_3 u_1 + v_2 - e_v & a_3 v_1 & a_3 \\
        a_4 u_1 & a_4 v_1 + v_2 - e_v & a_4 \\
        u_1 e_u - u_1 u_2 & v_1 e_u - v_1 u_2 & e_u - u_2 \\
        u_1 e_v - u_1 v_2 & v_1 e_v - v_1 v_2 & e_v - v_2
	\end{bmatrix},
\end{equation*}
where $e_u$ and $e_v$ are the coordinates of the epipole in the second image. The optimal solution in the least squares sense is $\matr{x} = \matr{C}^\dag \matr{b}$, where $\matr{C}^\dag = (\matr{C}^\trans \matr{C})^{-1} \matr{C}^\trans$ is the Moore-Penrose pseudo-inverse of $\matr{C}$. 

According to the experiments in~\cite{barath2017theory}, the method is often superior to the widely used solvers and makes robust estimation significantly faster due to the low number of points needed. However, its drawback is the necessity of the affine features which are time consuming to obtain in real world. By applying the algorithm proposed in this paper, it is possible to use the HAF method with SIFT features as input. Due to having real time SIFT implementations, e.g.\ \cite{sinha2006gpu}, the method is easy to be applied online. 
In this section, we test HAF getting its input by the proposed algorithm both in our synthesized environment and on publicly available real world datasets. 

\begin{figure*}[htb]
	\centering
    \begin{subfigure}[t]{0.328\textwidth}
        \includegraphics[width=0.99\textwidth]{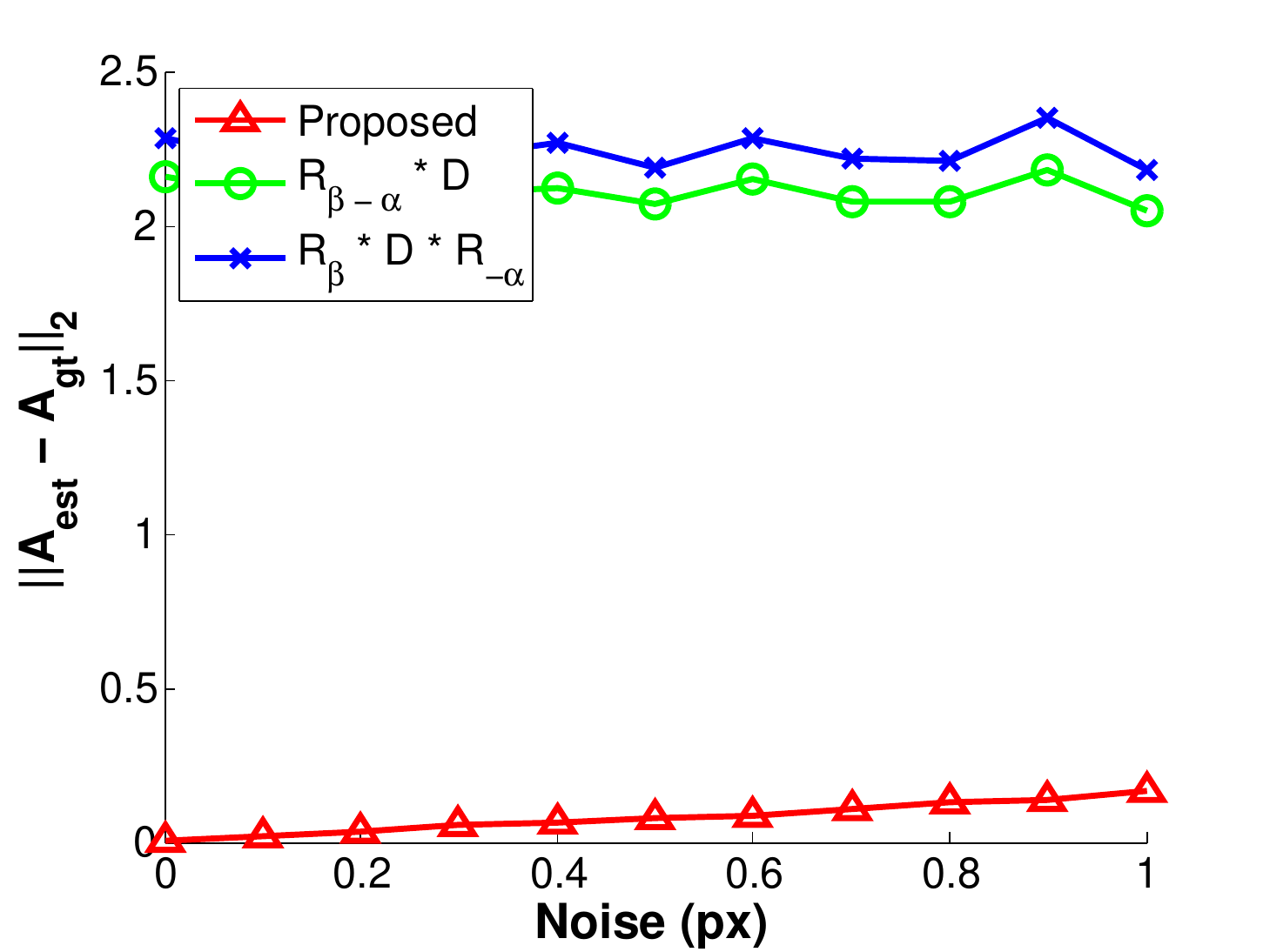}
        \caption{ }
        \label{fig:affine_error_increasing_noise}
    \end{subfigure}
    \begin{subfigure}[t]{0.328\textwidth}
        \includegraphics[width=0.99\textwidth]{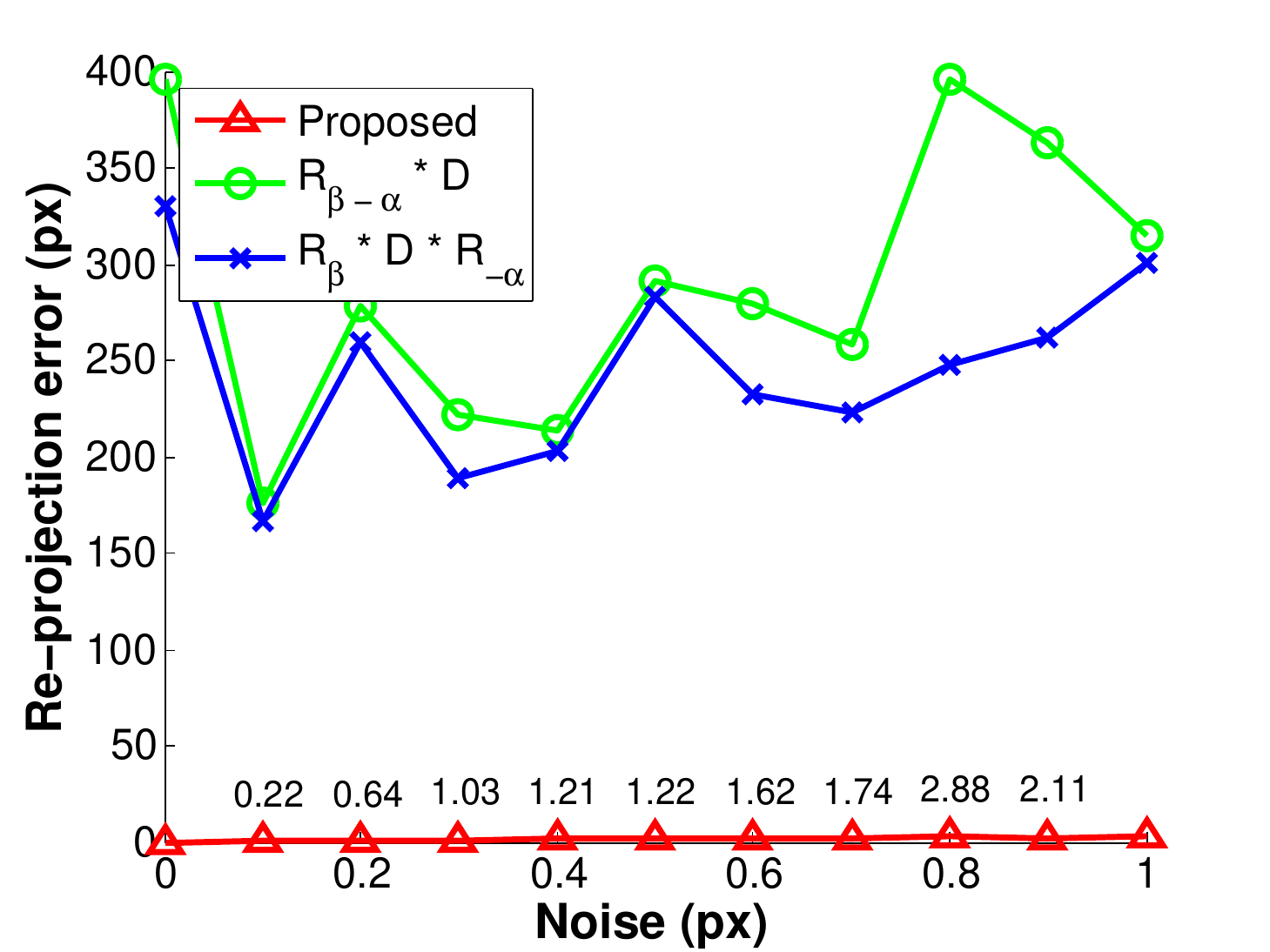}
        \caption{ }
        \label{fig:homography_comp_affines}
    \end{subfigure}
    \begin{subfigure}[t]{0.328\textwidth}
        \includegraphics[width=0.99\textwidth]{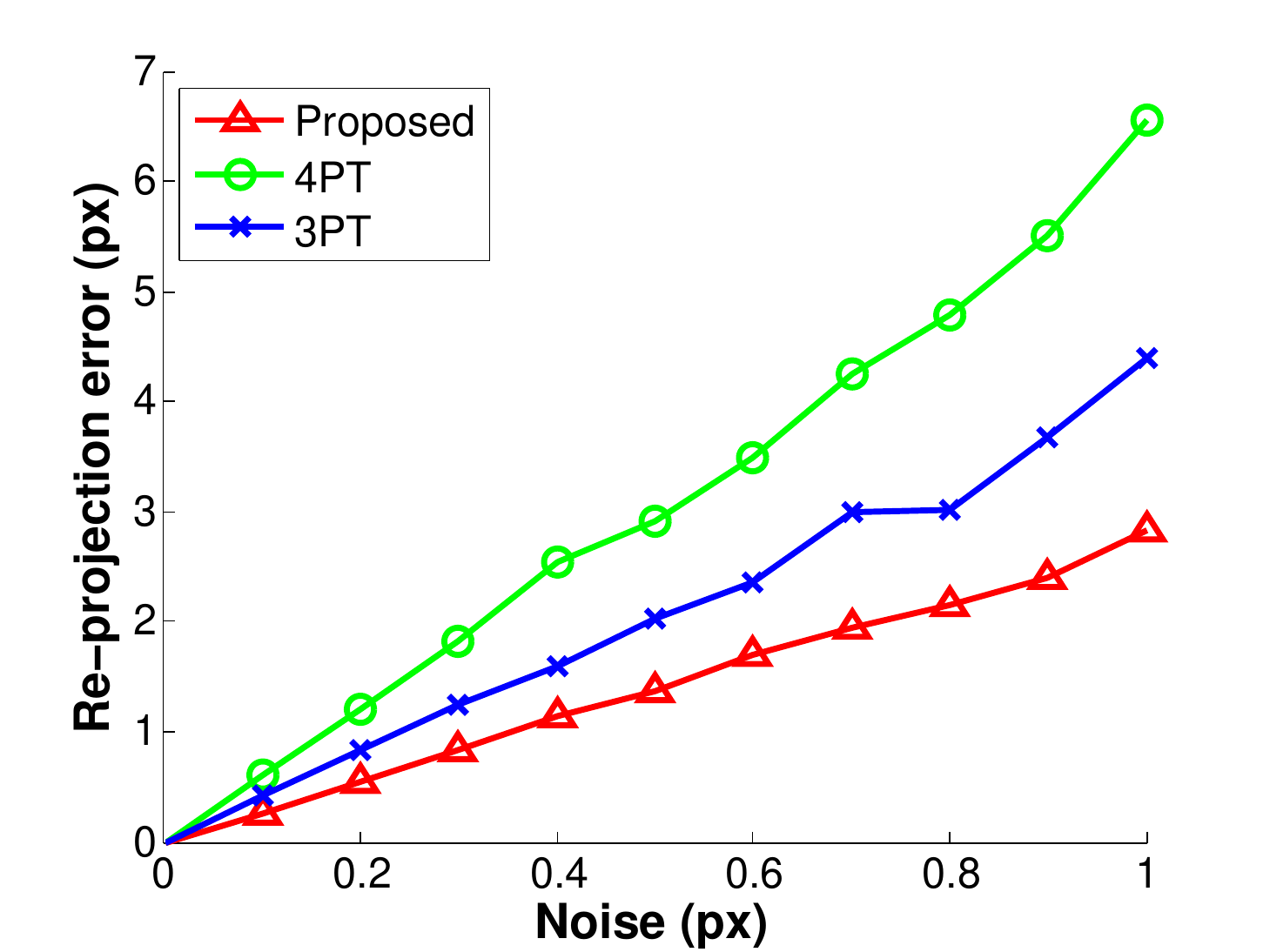}
        \caption{  }
        \label{fig:homography_comp_methods}
    \end{subfigure}
    \begin{subfigure}[t]{0.328\textwidth}
        \includegraphics[width=0.99\textwidth]{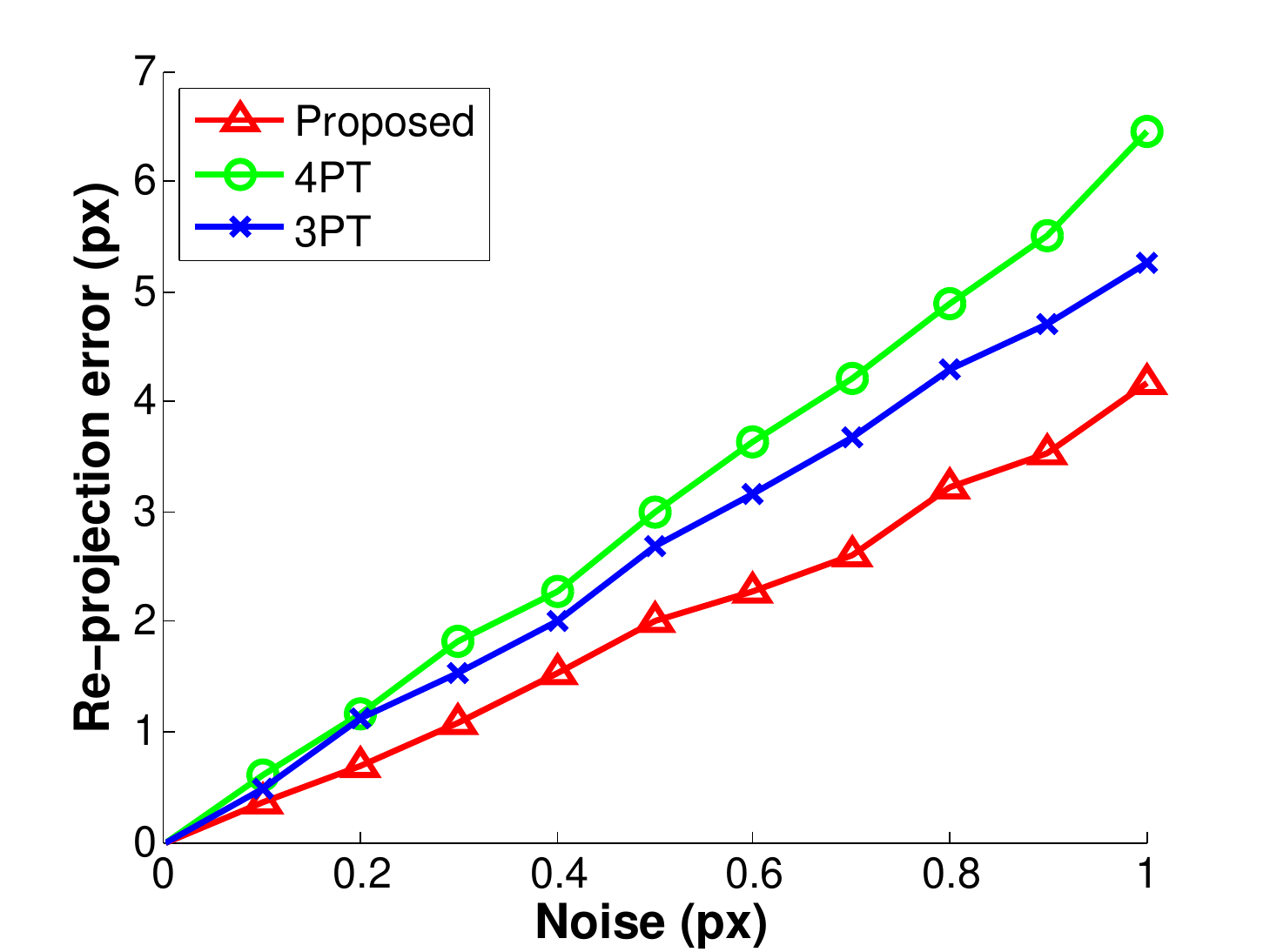}
        \caption{ }
        \label{fig:homography_comp_methods_noisy_F}
    \end{subfigure}
    \begin{subfigure}[t]{0.328\textwidth}
        \includegraphics[width=0.99\textwidth]{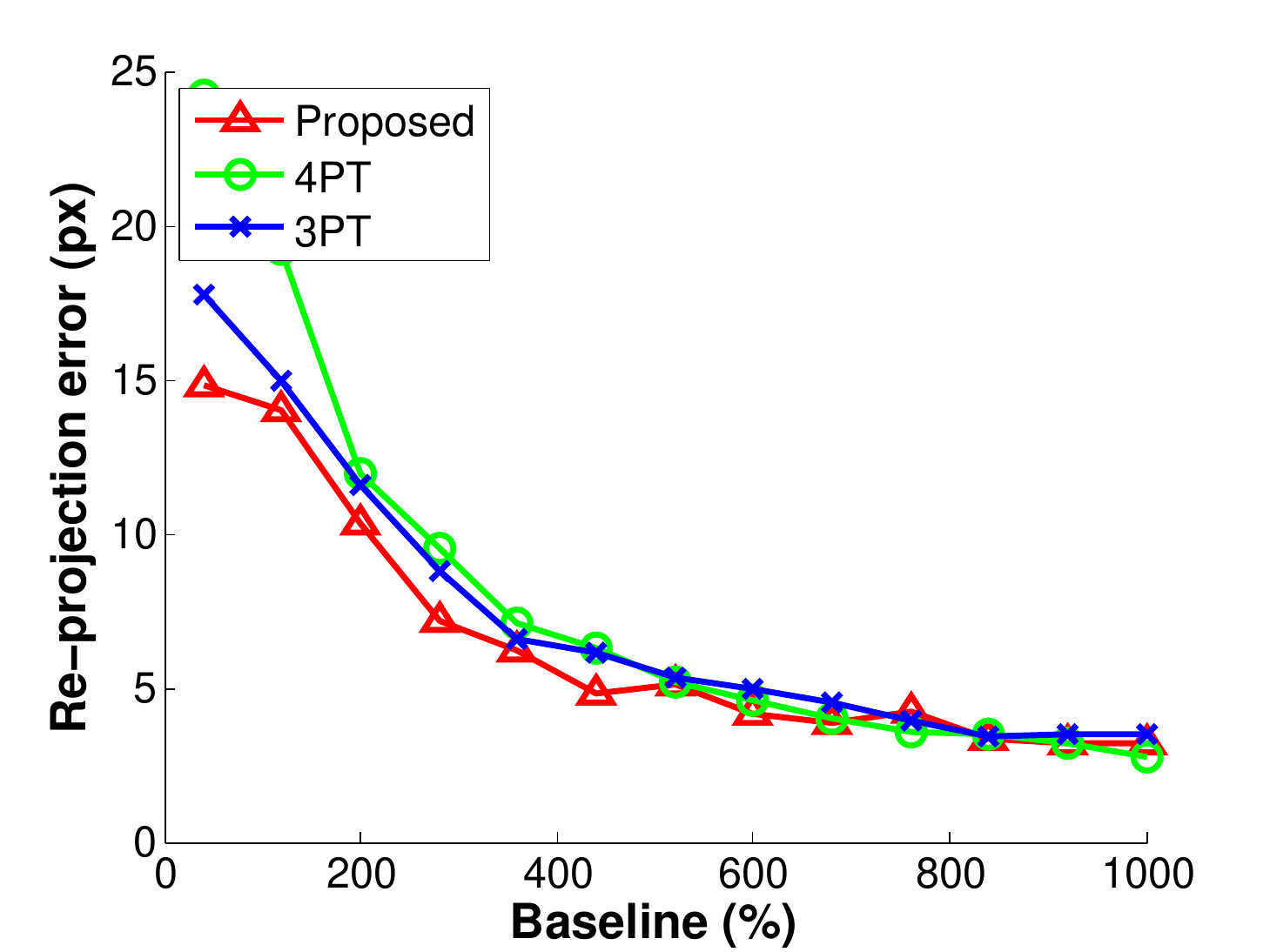}
        \caption{ }
        \label{fig:homography_comp_increasing_baseline}
    \end{subfigure}
    \caption{ (a) The error, i.e.\ $|| \Aff_{\text{est}} - \Aff_{\text{gt}} ||_\text{F}$, of the estimated affinities plotted as the function of the noise $\sigma$ added to the point coordinates. The affinities were recovered by the proposed method (red curve), approximated as $\textbf{A} \approx \matr{R}_{\beta - \alpha} \textbf{D}$ (green) and as $\textbf{A} \approx\matr{R}_{\beta} \textbf{D} \matr{R}_{-\alpha}$ (blue), where $\textbf{R}_\theta$ is 2D rotation by $\theta$ degrees and $\textbf{D} = \text{diag}(q, q)$. (\textbf{b--e}) Homography estimation from synthetic data. The horizontal axes show the noise $\sigma$ (px) added to the point coordinates. The vertical axes report the re-projection error (px; avg. of 1000 runs) computed from correspondences not used for the estimation. (b) Estimation by the HAF method~\cite{barath2017theory} from affine correspondences recovered in different ways. (c--e) Comparison of estimators: the proposed, the normalized four- (4PT) and three-point (3PT)~\cite{barath2017theory} algorithms. For (c), the ground truth fundamental matrix was used. For (d), it was estimated from the noisy correspondences by the normalized eight-point algorithm~\cite{hartley2003multiple}. For (d), the fundamental matrix was estimated, the noise $\sigma$ was $0.5$ pixels, and the errors are plotted as the function of the baseline between the cameras. The baseline is the ratio (\%) of the distances of the $1$st camera from the $2$nd one and from the origin. }
\end{figure*}

\begin{figure*}[htb]
	\centering
    \includegraphics[width=0.325\textwidth]{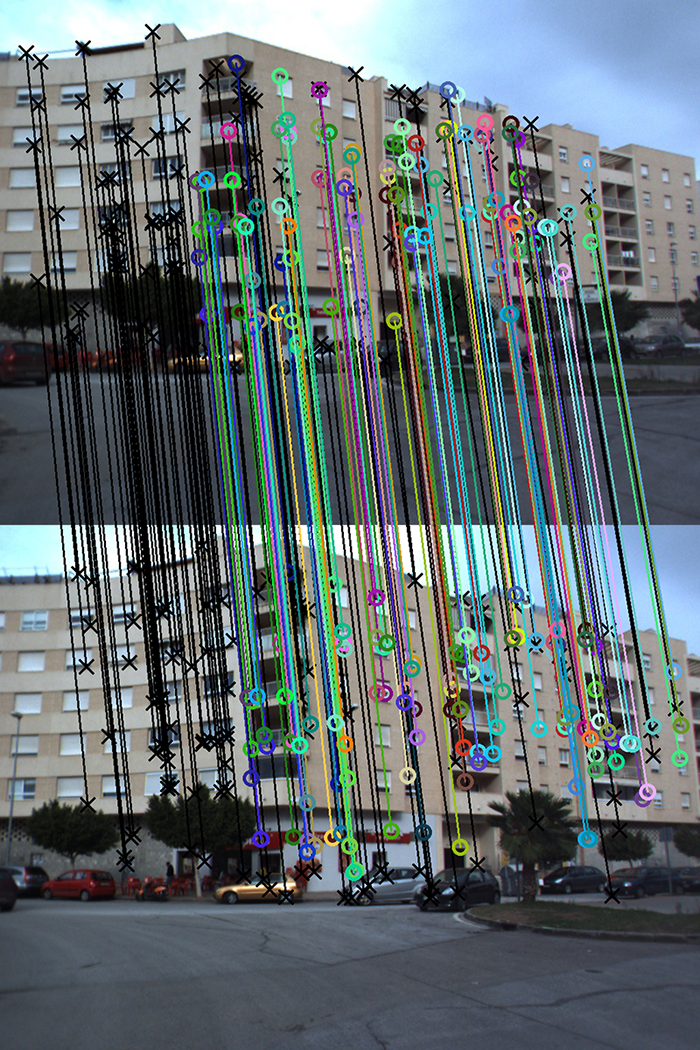}
    \includegraphics[width=0.325\textwidth]{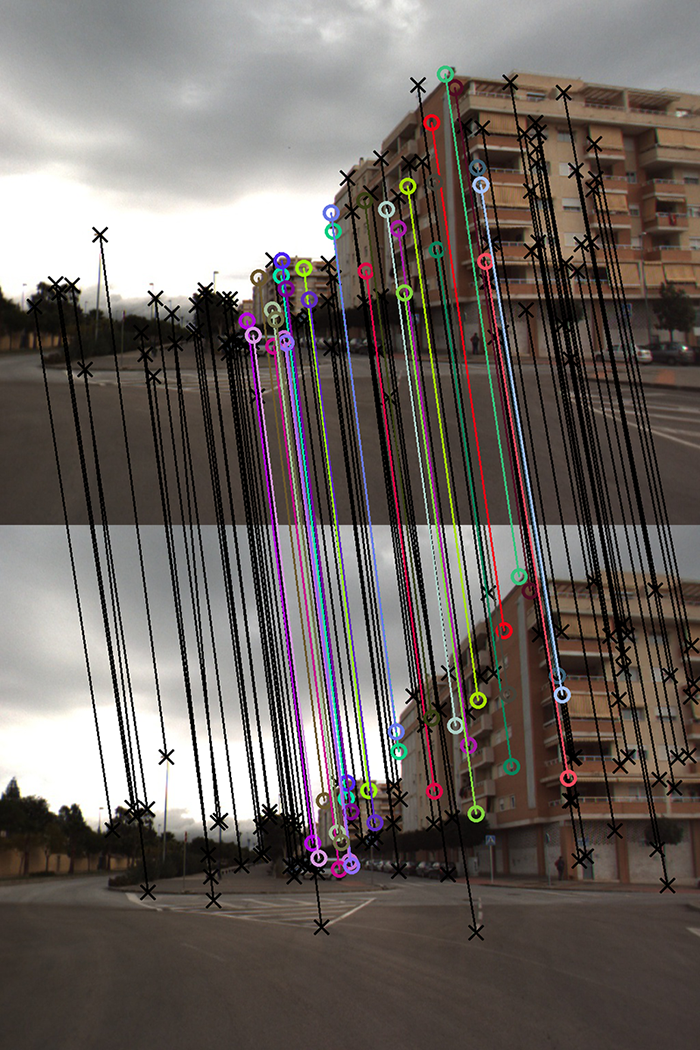}
    \includegraphics[width=0.325\textwidth]{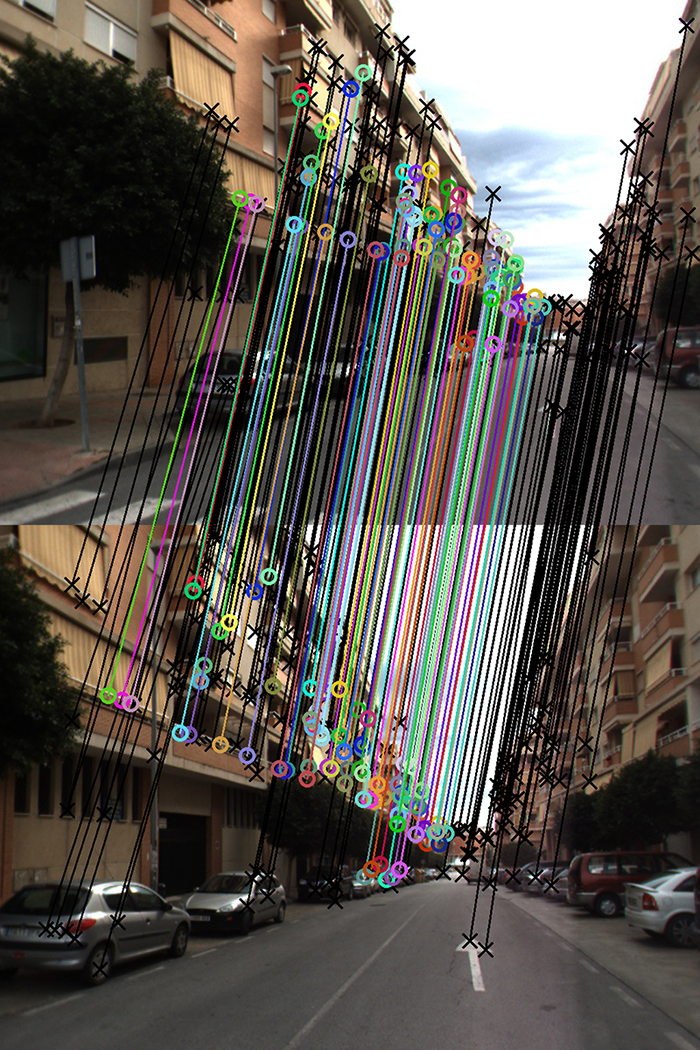}
    \caption{ Inlier (circles) and outlier (black crosses) correspondences found by the proposed method on image pairs from the {\fontfamily{cmtt}\selectfont Malaga} dataset. Every $5$th point is drawn. }
    \label{fig:datasets_2}
\end{figure*}

\begin{figure*}[htb]
	\centering
    \hspace*{-0.6cm}
    \includegraphics[width=1.08\textwidth]{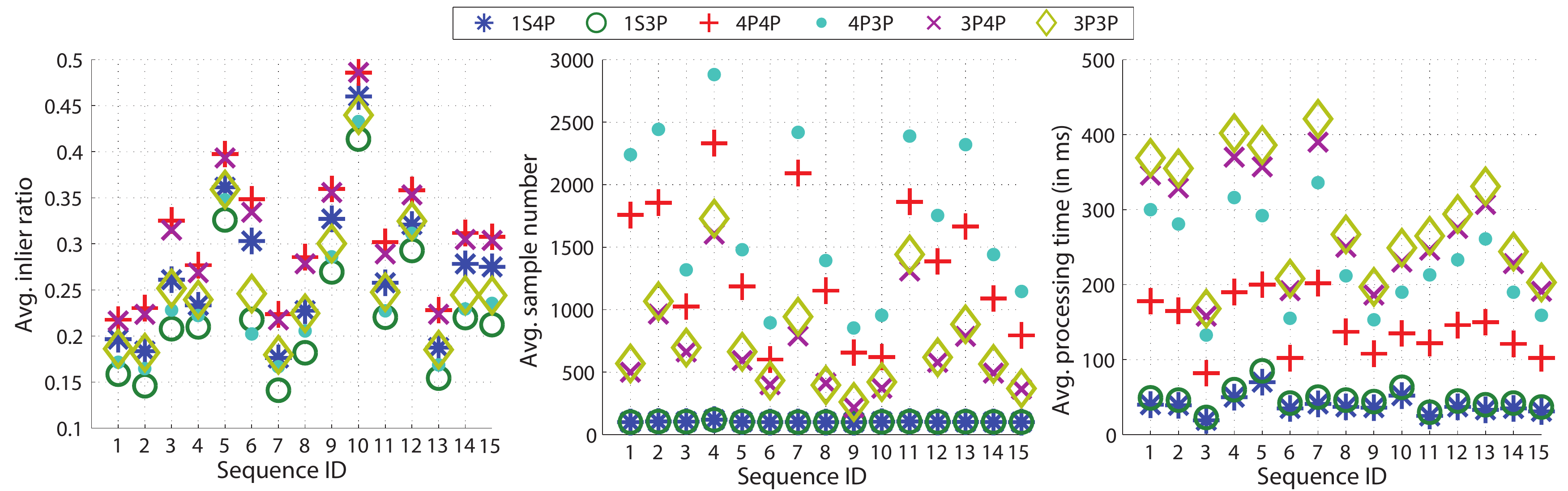}
    \caption{ Homography estimation on the {\fontfamily{cmtt}\selectfont Malaga} dataset. The horizontal axis shows the identifier of the image sequence. Each column reports the mean result on the image pairs. In total, $9064$ pairs were used. See Table~\ref{tab:hom_abbrevations} for the description of the methods. The shown properties are: the inlier ratio (left),  the number of samples drawn by LO-RANSAC (middle) and the processing time in milliseconds (right). }
    \label{fig:malaga_inliers}
\end{figure*}
 
\paragraph{Synthesized tests.} 
For testing the accuracy of homography estimation, we used the same synthetic scene as for the previous experiments. 
For Figure~\ref{fig:homography_comp_affines}, the homographies were estimated from affine correspondences recovered or approximated from the two rotations and the scale in different ways (similarly as in the previous section). 
Homography $\Hom$ was calculated from the recovered and also from the approximated affine features by the HAF method~\cite{barath2017theory}. In order to measure the accuracy of $\Hom$, ten random points were projected to the cameras from the 3D plane inducing $\Hom$ and the average re-projection error was calculated (vertical axis; average of $1000$ runs) and plotted as the function of the noise $\sigma$ (horizontal axis). 
To Figure~\ref{fig:homography_comp_affines}, these approximations lead to fairly rough results. The error is significant even in the noise-free case.
Also, it can be seen that the proposed method leads to perfect results in the noise-free case and the error behaves reasonably as the noise increases.  

In Figures~\ref{fig:homography_comp_methods} and \ref{fig:homography_comp_methods_noisy_F}, the HAF method with its input got from the proposed algorithm is compared with the normalized four-point~\cite{hartley2003multiple} (4PT) and three-point~\cite{barath2017theory} (3PT) methods. The re-projection error (vertical axis; average of $1000$ runs) is plotted as the function of the noise $\sigma$ (horizontal axis). For Figure~\ref{fig:homography_comp_methods}, the ground truth fundamental matrix was used. For Figure~\ref{fig:homography_comp_methods_noisy_F}, $\Fund$ was estimated from the noisy correspondences by the normalized eight-point algorithm~\cite{hartley2003multiple}. 
It can be seen that the HAF algorithm using the calculated affinities as input leads to the most accurate homographies in both cases.

Figure~\ref{fig:homography_comp_increasing_baseline} shows the sensitivity to the baseline (horizontal axis; in percentage). The baseline is considered as the ratio of the distance of the two cameras and the distance of the first one from the observed point. It can be seen that all methods are fairly sensitive to the baseline and obtain instable results for image pairs with short one. However, HAF method is slightly more accurate than the other competitors. 

\paragraph{Real world tests.} 
In order to test the proposed method on real world data, we used the 
{\fontfamily{cmtt}\selectfont AdelaideRMF}\footnote{{cs.adelaide.edu.au/~hwong/doku.php?id=data}}, 
{\fontfamily{cmtt}\selectfont Multi-H}\footnote{{web.eee.sztaki.hu/~dbarath}} and {\fontfamily{cmtt}\selectfont Malaga}\footnote{{www.mrpt.org/MalagaUrbanDataset}} 
datasets (see Figures~\ref{fig:datasets_1} and \ref{fig:datasets_2} for example image pairs). 
{\fontfamily{cmtt}\selectfont AdelaideRMF} and {\fontfamily{cmtt}\selectfont Multi-H} consist of image pairs of resolution from $455 \times 341$ to $2592 \times 1944$ and manually annotated (assigned to a homography, i.e.\ a plane, or to the outlier class) correspondences. Since the reference point sets do not contain rotations and scales, we detected and matched points applying the SIFT detector. 
Ground truth homographies were estimated from the manually annotated correspondences. For each of them, we selected the points out of the detected SIFT correspondences which are closer than a manually set inlier-outlier threshold, i.e.\ $2$ pixels. 
The {\fontfamily{cmtt}\selectfont Malaga} dataset was gathered entirely in urban scenarios with a car equipped with several sensors, including a high-resolution camera and five laser scanners. We used the $15$ video sequences taken by the camera and every $10$th image from each sequence. 
As a robust estimator, we chose LO-RANSAC~\cite{chum2003locally} with PROSAC~\cite{chum2005matching} sampling and inlier-outlier threshold set to $2$ pixels. All the real solutions of the proposed method were considered and validated as different homographies against the input correspondences. 

Given an image pair, the procedure to evaluate the estimators, i.e.\ the proposed, the four- and three-point algorithms, is as follows: 
\begin{enumerate}
\item A fundamental matrix was estimated by LO-RANSAC using the seven-point algorithm as a minimal method, the normalized eight-point algorithm for least-squares fitting, the Sampson-distance as residual function, and a threshold set to $2$ pixels. 
\item The ground truth homographies, estimated from the manually annotated correspondence sets, were selected one by one. For each homography:
	\begin{enumerate}
    	\item The correspondences which did not belong to the selected homography were replaced by completely random correspondences to reduce to probability of finding a different plane than what was currently tested. 
        \item LO-RANSAC combined with the compared estimators was applied to the point set consisting of the inliers of the current homography and outliers. 
        \item The estimated homography is compared against the ground truth one estimated from the manually selected inliers.
    \end{enumerate}
\end{enumerate}
In LO-RANSAC (step 2b), there are two cases when a model is estimated: (i) when a minimal sample is selected and (ii) during the local optimization by fitting to a set of inliers. 
To select the estimators used for these steps, there is a number of possibilities exists. For instance, it is possible to use the proposed algorithm for fitting to a minimal sample and the normalized four-point algorithm for the least-squares fitting. 
The tested combinations are reported in Table~\ref{tab:hom_abbrevations}.
\begin{table}[ht]
\center
	\caption{ The compared settings for locally optimized RANSAC~\cite{chum2003locally}. The first column shows the abbreviations of the combinations, the second one contains the methods applied for fitting a homography to a minimal sample and the last one consists of the algorithms used for the least-squares fitting. The two light gray rows are the ones which use the proposed method. }
  	\begin{tabular}{| c | c | c | }
    \hline 
         \multicolumn{3}{ | c | }{ \cellcolor{black!20}{Abbreviations}\rule{0pt}{2.0ex} }  \\
    \hline 
         \phantom{xx}\textbf{Name}\phantom{xx}\rule{0pt}{2.0ex} & \phantom{xx}\textbf{Minimal method}\phantom{xx}\rule{0pt}{2.0ex} & \phantom{xx}\textbf{Least-squares fitting}\phantom{xx}\rule{0pt}{2.0ex} \\
    \hline 
         \cellcolor{black!10}1S4P & \cellcolor{black!10}Proposed algorithm & \cellcolor{black!10}Four-point method \\
         \cellcolor{black!10}1S3P & \cellcolor{black!10}Proposed algorithm & \cellcolor{black!10}Three-point method \\
         4P4P & Four-point method & Four-point method \\
         4P3P & Four-point method & Three-point method \\
         3P4P & Three-point method & Four-point method \\
         3P3P & Three-point method & Three-point method \\
    \hline 
	\end{tabular}
	\label{tab:hom_abbrevations}
\end{table}

Table~\ref{tab:dataset_comparison} reports the results on the {\fontfamily{cmtt}\selectfont AdelaideRMF} and {\fontfamily{cmtt}\selectfont Multi-H} datasets (first column). The number of planes in each dataset is written into brackets. The tested methods are shown in the second column (see Table~\ref{tab:hom_abbrevations}). The blocks, each consisting of four columns, show the percentage of the homographies not found (FN, in \%); the mean re-projection error of the found ones computed from the manually annotated correspondences and the estimated homographies ($\epsilon$, in pixels); the number of samples drawn by LO-RANSAC (s); and the processing time (t, in milliseconds). We considered a homography as a not found one if the re-projection error was higher than $10$ pixels. For the first block, the required confidence of LO-RANSAC in the results was set to $0.95$, for the second one, it was $0.99$. The values were computed as the means of $100$ runs on each homography selected. The methods which applied the proposed technique are 1S4P and 1S3P.
It can be seen that the proposed algorithm, in terms of accuracy and number of planes found, leads to similar results to that of the competitor algorithms -- sometimes worse sometimes more accurate. However, due to requiring only a sole correspondence, both its processing time and number of samples required are almost an order of magnitude lower than that of the other methods.

The results on the {\fontfamily{cmtt}\selectfont Malaga} dataset are shown in Figure~\ref{fig:malaga_inliers}. In total, $9064$ image pairs were tested. The confidence of LO-RANSAC was set to $0.99$ and the inlier-outlier threshold to $2.0$ pixels. The reported properties are the ratio of the inliers found (left plot; vertical axis), the number of samples drawn by LO-RANSAC (middle; vertical axis) and the processing time in milliseconds (right; vertical axis). The horizontal axes show the identifiers of the image sequences in the dataset. 
The same trend can be seen as for the other datasets. The ratio of inliers found is similar to that of the competitor algorithms. The number of samples used and the processing time is significantly lower than that of the other techniques. 

\begin{table*}[ht]
\center
\caption{ Homography estimation on the {\fontfamily{cmtt}\selectfont AdelaideRMF} ($18$ pairs; $43$ planes; from the $3$rd to the $8$th rows) and {\fontfamily{cmtt}\selectfont Multi-H} ($4$ pairs; $33$ planes; from $9$th to $14$th rows) datasets by LO-RANSAC~\cite{chum2003locally} combined with minimal methods. Each row reports the results of a method (gray color indicates the proposed one). See Table~\ref{tab:hom_abbrevations} for the abbreviations. The required confidence of LO-RANSAC was set to $0.95$ for the $3$--$6$th and to $0.99$ for the $7$--$10$th columns. The reported properties are: the ratio of homographies not found (FN, in percentage); the mean re-projection error ($\epsilon$, in pixels); the number of samples drawn by LO-RANSAC (s); and the processing time (t, in milliseconds). Average of $100$ runs.}
  	\begin{tabular}{| c | c || c c c c | c c c c | }
    \hline 
 	 	 \mc{2}{| c}{\cellcolor{black!20}} & \mc{4}{| c}{\cellcolor{black!20}\ph{xx}Confidence 95\%\ph{xx}} & \mc{4}{c |}{\cellcolor{black!20}\ph{xx}Confidence 99\%\ph{xx}} \\ 
    \hline    
 	 	 \ph{xxx} & \mc{1}{c}{} & \ph{xxx}FN\ph{xxx} & \ph{xxx}$\epsilon$\ph{xxx} & \ph{xxx}s\ph{xxx} & \ph{xxx}t\ph{xx} & \ph{xxx}FN\ph{xxx} & \ph{xxx}$\epsilon$\ph{xxx} & \ph{xxx}s\ph{xxx} & \ph{xxx}t\ph{xxx} \\ 
    \hline
         \multirow{6}{*}{\rot{Adelaide ($43$\#)}} & \cellcolor{Gray}1S4P & \cellcolor{Gray}2.28 & \cellcolor{Gray}1.62 & \cellcolor{Gray}\ph{x}125 &  \cellcolor{Gray}168 & \cellcolor{Gray}2.15 & \cellcolor{Gray}1.60 & \cellcolor{Gray}\ph{x}147 & \cellcolor{Gray}174\\
         & \cellcolor{Gray}1S3P & \cellcolor{Gray}1.09 & \cellcolor{Gray}1.60 & \cellcolor{Gray}\ph{x}\textbf{118} &  \cellcolor{Gray}\ph{x}\textbf{67} & \cellcolor{Gray}0.86 & \cellcolor{Gray}1.60 & \cellcolor{Gray}\ph{x}\textbf{129} & \cellcolor{Gray}\ph{x}\textbf{68} \\
         & 4P4P & 1.39 & 1.56 & 2172 & 321 & 1.17 & \textbf{1.57} & 2351 & 338\\
         & 4P3P & 0.62 & \textbf{1.57} & 1928 & 173 & 0.72 & 1.59 & 2179 & 188\\
         & 3P4P & \textbf{0.08} & 1.60 & \ph{x}977 &  250 & \textbf{0.12} & 1.60 & 1224 & 260\\
         & \ph{x}3P3P\ph{x} & 0.39 & 1.63 & 1070 & 107 & 0.51 & 1.62 & 1278 & 113\\
    \hline
         \multirow{6}{*}{\rot{Multi-H ($33$\#)}} & \cellcolor{Gray}1S4P & \cellcolor{Gray}11.31 & \cellcolor{Gray}1.91 & \cellcolor{Gray}1872 & \cellcolor{Gray}222 & \cellcolor{Gray}10.13 & \cellcolor{Gray}\textbf{1.55} & \cellcolor{Gray}1993 & \cellcolor{Gray}250 \\
         & \cellcolor{Gray}1S3P & \cellcolor{Gray}\ph{x}9.14 & \cellcolor{Gray}\textbf{1.89} & \cellcolor{Gray}\textbf{1816} & \cellcolor{Gray}\textbf{198} & \cellcolor{Gray}\ph{x}8.26 & \cellcolor{Gray}1.57 & \cellcolor{Gray}\textbf{1936} & \cellcolor{Gray}\textbf{218} \\
         & 4P4P & 12.10 & 2.73 & 4528 & 538 & 12.40 & 2.68 & 4581 & 528 \\
         & 4P3P & 10.30 & 2.46 & 4069 & 483 & 10.13 & 2.29 & 4103 & 480 \\
         & 3P4P & 10.05 & 2.33 & 4312 & 447 & \ph{x}9.09 & 2.09 & 4244 & 450 \\
         & 3P3P & \ph{x}\textbf{8.74} & 1.94 & 4105 & 385 & \ph{x}\textbf{8.11} & 1.80 & 4110 & 377 \\
    \hline    
\end{tabular}
\label{tab:dataset_comparison}
\end{table*}

\section{Conclusion}

An approach is proposed for recovering affine correspondences from orientation- and scale-invariant features obtained by, for instance, SIFT or SURF detectors.  The method estimates the affine correspondence by enforcing the geometric constraints which the pre-estimated epipolar geometry implies. The solution is obtained in closed-form as the roots of a quadratic polynomial equation. Thus the estimation is extremely fast, i.e.\ $<1$ millisecond, and leads to at most two real solutions. It is demonstrated on synthetic and publicly available real world datasets -- containing more than $9000$ image pairs -- that by using the proposed technique correspondence-wise homography estimation is possible. The geometric accuracy of the obtained homographies is similar to that of the state-of-the-art algorithms. However, due to requiring only a single correspondence, the robust estimation, e.g.\ by LO-RANSAC, is an order of magnitude faster than by using the four- or three-point algorithms.

\appendix

\section*{Affine Decomposition}

The decomposition of local affine transformation $\Aff$ to two rotations ($\matr{R}_\gamma \in \mathbb{R}^{2 \times 2}$ and $\matr{R}_\delta \in \mathbb{R}^{2 \times 2}$) and an upper triangle matrix ($\matr{U} \in \mathbb{R}^{2 \times 2}$) is discussed in this section. The problem is as follows:
\begin{equation*}
	\Aff = \matr{R}_\gamma \matr{U} \matr{R}_\delta.
    \label{eq:decomposition}
\end{equation*}
Note that we write the decomposition generally, thus the angles, denoted by $\gamma$ and $\delta$, does not correspond to the orientation of any features. This is the reason why we do not write $\matr{R}_{-\alpha_1}$ instead of $\matr{R}_\delta$. After multiplying the three matrices, the following system is given for the affine components:
\begin{eqnarray}
	\begin{array}{ccc}
      a_1 & = & c_\gamma c_\delta q_u + c_\gamma s_\delta w - s_\gamma q_v s_\delta, \\[1mm]
      a_2 & = & c_\gamma c_\delta w - c_\gamma s_\delta q_u - s_\gamma c_\delta q_v, \\[1mm]
      a_3 & = & s_\gamma c_\delta q_u + s_\gamma s_\delta w + c_\gamma s_\delta q_v, \\[1mm]
      a_4 & = & s_\gamma c_\delta w - s_\gamma s_\delta q_u + c_\gamma c_\delta q_v, 
	\end{array}
    \label{eq:decomposition_system}
\end{eqnarray}
where $c_\gamma = \cos(\gamma)$, $s_\gamma = \sin(\gamma)$, $c_\delta = \cos(\delta)$, $s_\delta = \sin(\delta)$; scalars $q_u$ and $q_v$ are the scales along axes $x$ and $y$; and $w$ is the shear. Since there are four equations for five unknowns, the decomposition is not unique. Considering that in real world setting, the used features are usually orientation-invariant ones, we chose angle $\gamma \in [0, 2\pi)$ to parameterize the possible decompositions. Thus, for each $\gamma$, there will be a unique decomposition of $\Aff$. 

Solving Eqs.~\ref{eq:decomposition_system} lead to two possible $\delta$s, each providing a valid solution, as follows:
\begin{eqnarray*}
	\delta_{12} = \cos^{-1} (\pm (c_\gamma a_4 - s_\gamma a_2)\\[2mm]
    \frac{ \sqrt{c_\gamma^2 (a_4^2 + a_3^2) - 2 c_\gamma s_\gamma (a_2 a_4 + a_1 a_3) + s_\gamma^2 (a_2^2 + a_1^2)}}{c_\gamma^2 (a_4^2 + a_3^2) - 2 c_\gamma s_\gamma (a_2 a_4 + a_1 a_3) + s_\gamma^2 (a_2^2 + a_1^2)})
\end{eqnarray*}
Since both $\delta$s are valid, the computation of the remaining affine components falls apart to two cases -- they have to be calculated for each $\delta$ independently. The formulas for the scales ($q_u$ and $q_v$) and the shear ($w$) are as follows: 
\begin{eqnarray*}
  	\begin{array}{l}
        q_u = -\frac{a_\delta s_\delta - a_\gamma c_\delta}{c_\gamma s_\delta^2 + c_\gamma c_\delta^2} \\[3mm]
        q_v = -\frac{s_\gamma a_\gamma - c_\gamma a_3}{(s_\gamma^2 + c_\gamma^2) s_\delta} \\[3mm]
        w = \frac{(c_\gamma s_\gamma a_3 + c_\gamma^2 a_\gamma) s_\delta^2 + (s_\gamma^2 a_\delta + c_\gamma^2 a_\delta) c_\delta s_\delta + (c_\gamma s_\gamma a_3 - s_\gamma^2 a_\gamma) c_\delta^2}{(c_\gamma s_\gamma^2+c_\gamma^3) s_\delta^3 + (c_\gamma s_\gamma^2 + c_\gamma^3) c_\delta^2 s_\delta},
  	\end{array}
\end{eqnarray*}
where $\delta \in \{\delta_1, \delta_2\}$. Finally, the decomposition is selected for which $||\Aff - \matr{R}_\gamma^i \matr{U}^i \matr{R}_\delta^i ||_2$ is minimal ($i \in \{1,2\}$).

{\small
\bibliographystyle{splncs}
\bibliography{egbib}

\begin{thebibliography}{10}

\bibitem{mikolajczyk2005comparison}
Mikolajczyk, K., Tuytelaars, T., Schmid, C., Zisserman, A., Matas, J.,
  Schaffalitzky, F., Kadir, T., Van~Gool, L.:
\newblock A comparison of affine region detectors.
\newblock International journal of computer vision \textbf{65} (2005)  43--72

\bibitem{lowe1999object}
Lowe, D.G.:
\newblock Object recognition from local scale-invariant features.
\newblock In: International Conference on Computer vision. (1999)

\bibitem{bay2006surf}
Bay, H., Tuytelaars, T., Van~Gool, L.:
\newblock {SURF}: Speeded up robust features.
\newblock European Conference on Computer Vision (2006)

\bibitem{PerdochMC06}
Perdoch, M., Matas, J., Chum, O.:
\newblock Epipolar geometry from two correspondences.
\newblock In: International Conference on Pattern Recognition. (2006)

\bibitem{Bentolila2014}
Bentolila, J., Francos, J.M.:
\newblock Conic epipolar constraints from affine correspondences.
\newblock Computer Vision and Image Understanding (2014)

\bibitem{Raposo2016}
Raposo, C., Barreto, J.P.:
\newblock Theory and practice of structure-from-motion using affine
  correspondences.
\newblock In: Computer Vision and Pattern Recognition. (2016)

\bibitem{barath2017focal}
Barath, D., Toth, T., Hajder, L.:
\newblock A minimal solution for two-view focal-length estimation using two
  affine correspondences.
\newblock In: Conference on Computer Vision and Pattern Recognition. (2017)

\bibitem{koser2009geometric}
K{\"o}ser, K.:
\newblock Geometric Estimation with Local Affine Frames and Free-form Surfaces.
\newblock Shaker (2009)

\bibitem{barath2017theory}
Barath, D., Hajder, L.:
\newblock A theory of point-wise homography estimation.
\newblock Pattern Recognition Letters \textbf{94} (2017)  7--14

\bibitem{Molnar2014}
Moln\'ar, J., Chetverikov, D.:
\newblock Quadratic transformation for planar mapping of implicit surfaces.
\newblock Journal of Mathematical Imaging and Vision (2014)

\bibitem{Pritts2017RadiallyDistortedCT}
Pritts, J., Kukelova, Z., Larsson, V., Chum, O.:
\newblock Radially-distorted conjugate translations.
\newblock Conference on Computer Vision and Pattern Recognition (2018)

\bibitem{morel2009asift}
Morel, J.M., Yu, G.:
\newblock {ASIFT}: A new framework for fully affine invariant image comparison.
\newblock SIAM journal on imaging sciences \textbf{2} (2009)  438--469

\bibitem{mishkin2015mods}
Mishkin, D., Matas, J., Perdoch, M.:
\newblock {MODS}: Fast and robust method for two-view matching.
\newblock Computer Vision and Image Understanding (2015)

\bibitem{matas2004robust}
Matas, J., Chum, O., Urban, M., Pajdla, T.:
\newblock Robust wide-baseline stereo from maximally stable extremal regions.
\newblock Image and vision computing (2004)

\bibitem{barath2017phaf}
Barath, D.:
\newblock {P-HAF}: Homography estimation using partial local affine frames.
\newblock In: International Conference on Computer Vision Theory and
  Applications. (2017)

\bibitem{barath2018five}
Barath, D.:
\newblock Five-point fundamental matrix estimation for uncalibrated cameras.
\newblock Conference on Computer Vision and Pattern Recognition (2018)

\bibitem{chum2003locally}
Chum, O., Matas, J., Kittler, J.:
\newblock Locally optimized ransac.
\newblock In: Joint Pattern Recognition Symposium. (2003)

\bibitem{hartley2003multiple}
Hartley, R., Zisserman, A.:
\newblock Multiple view geometry in computer vision.
\newblock Cambridge University Press (2003)

\bibitem{sinha2006gpu}
Sinha, S.N., Frahm, J.M., Pollefeys, M., Genc, Y.:
\newblock Gpu-based video feature tracking and matching.
\newblock In: Workshop on Edge Computing Using New Commodity Architectures.
  Volume 278. (2006)  4321

\bibitem{chum2005matching}
Chum, O., Matas, J.:
\newblock Matching with {PROSAC}-progressive sample consensus.
\newblock In: Computer Vision and Pattern Recognition. (2005)

\end{thebibliography}
}

\end{document}